\documentclass{article}
\usepackage{spconf,amsmath,graphicx}
\usepackage{multirow}
\usepackage{url}
\usepackage{color}
\usepackage{hyperref}
\ninept

\title{Image denoising via group sparsity residual constraint}
\ninept
\name{\footnotesize{Zhiyuan Zha$^{1,3}$,  Xin Liu$^{2}$,   Ziheng Zhou$^{2}$, Xiaohua Huang $^{2}$, Jingang Shi $^{2}$, Zhenhong Shang $^{3}$, Lan Tang $^{1}$, Yechao Bai $^{1}$, Qiong Wang $^{1}$, Xinggan Zhang $^{1, \star}$}
\thanks{${\star}$ Corresponding Author. This work was supported by the NSFC (61571220, 61462052, 61502226) and nd the open research fund of National Mobile Commune. Research Lab., Southeast University (No.2015D08).}  }

\address{$^{1}$ \footnotesize{School of Electronic Science and Engineering, Nanjing University, Nanjing 210023, China.} \\
    $^{2}$ \footnotesize{The Center for Machine Vision and Signal Analysis, University of Oulu, 90014, Finland.}\\
    $^{3}$ \footnotesize{School of Information Engineering and Automation, Kunming University of Science and Technology, Kunming 650500, China.}}
\begin{document}
%
\maketitle
\begin{abstract}
Group sparsity has shown great potential in various low-level vision tasks (e.g, image denoising, deblurring and inpainting). In this paper, we propose a new prior model for image denoising via group sparsity residual constraint (GSRC).  To enhance the performance of group sparse-based image denoising, the concept of group sparsity residual is proposed, and thus, the problem of image denoising is translated into one that reduces the group sparsity residual. To reduce the residual, we first obtain some good estimation of the group sparse coefficients of the original image by the first-pass estimation of noisy image, and then centralize the group sparse coefficients of noisy image to the estimation. Experimental results have demonstrated that the proposed method not only outperforms many state-of-the-art denoising methods such as BM3D and WNNM, but results in a competitive speed.
\end{abstract}
\section{Introduction}
\label{sec:intro}
As a classical problem in low level vision, image denoising has been widely studied over the last half century due to its practical significance. Image denoising aims to estimate the  clean image $\textbf{\emph{X}}$ from its noisy observation $\textbf{\emph{Y}}=\textbf{\emph{X}}+\textbf{\emph{V}}$, where $\textbf{\emph{V}}$ is usually assumed to be additive white Gaussian noise. In the past decades, extensive studies have been conducted on developing various methods for image denoising. Due to the ill-posed nature of image denoising, it has been widely recognized that the prior knowledge of images plays a key role in enhancing the performance of image denoising methods. A variety of image prior models have been developed, including wavelet/curvelet based \cite{1,2}, total variation based \cite{3,4}, sparse representation based \cite{5,6,25} and nonlocal self-similarity based ones \cite{7,8}, etc.

Motivated by the fact that wavelet transform coefficients are actually regarded as Laplacian distribution, many wavelet shrinkage based methods have been proposed \cite{1,2}. For instance, Chang $\emph{et al}.$ \cite{1} proposed a method called Bayes shrink algorithm to model the wavelet transform coefficients as generalized Gaussian distribution. Remenyi $\emph{et al}.$ \cite{2} proposed to use 2D scale mixing complex-valued wavelet transform and achieved promising denoising performance. It has been acknowledged that natural image gradients have a heavy-tailed distribution. The image gradient is modeled as Laplacian distribution in the total variation based methods for image denoising \cite{3}.

Instead of modeling image statistics in some transform domain (e.g., gradient domain, wavelet domain, etc.), sparse representation based prior assumes that each patch of an image can be precisely represented by a sparse coefficient vector whose entries are mostly zero or close to zero based on a basis set known as a dictionary. The dictionary is usually learned from a natural image dataset \cite{5}. Compared with the conventional analytically designed dictionaries, such as those based on wavelet, curvelet and DCT, dictionaries learned directly from images have an advantage of being better adapted to local image structures \cite{5} and thus, could improve the sparsity which leads to the better denoising performance. Nonetheless, patch-based sparse representation model of natural images usually suffers from some limits, such as dictionary learning with great computational complexity, ignoring the relationship among similar patches.

Motivated by the fact that the image patches that have similar patterns can be spatially far from each other and thus can be gathered in the whole image, this so-called nonlocal self-similarity (NSS) prior is among the most remarkable priors for image restoration. The seminal work of nonlocal mean (NLM) \cite{7}, which initially utilized the NSS property to implement a form of the weighted filtering for image denoising. After this, inspired by the success of the NLM denoising filter, a flurry of nonlocal regularization methods\cite{6,9,10} were proposed to solve various image inverse problems. By contrast with the local regularization based methods, nonlocal regularization based methods can effectively generate sharper image edges and preserve more image details. However, there are still lots of image details and structures that cannot be accurately recovered. One important reason is that the above nonlocal regularization terms rely on the weighted graph \cite{11}, yet it is unavoidable that the weighted manner leads to disturbance and inaccuracy \cite{12}.

Based on the NSS property of an image, recent studies \cite{8,13,14,15} have revealed that structured or group sparsity can provide more powerful reconstruction performance for noise removal. For instance, Dabov $\emph{et al}.$ \cite{13} proposed BM3D method to exploit nonlocal similar patches and 3D transform domain collaborative filtering, which can achieve state-of-the-art performance in denoising. Marial $\emph{et al}.$ \cite{8} further advanced the idea of NSS by group sparse coding. As the matrix formed by nonlocal similar patches in a natural image is of a low rank, the low-rank modeling based methods \cite{14,15} have also achieved highly competitive denoising results.

Though group sparsity has verified its great success in image denoising,  only the NSS prior of noisy input image is used among a majority of existing methods for noise removal. For example, BM3D \cite{13} extracted the nonlocal similar patches from a noisy image and conducted collaborative filtering in the sparse 3D transform domain. In WNNM \cite{15}, the low-rank regularization is enforced to reconstruct the latent structure of the matrix of noisy patches. However, only considering the NSS property of noisy input images, it is very challenging to recover the latent clean image ${\textbf{\emph{X}}}$ directly from noisy observation $\textbf{\emph{Y}}$.

With the above question kept in mind, in this work, we propose a new prior model for image denoising by group sparse residual constraint (GSRC). To improve the performance of group sparse-based image denoising, the concept of group sparsity residual is proposed, and thus, the problem of image denoising is turned into one that reduces the group sparsity residual. To reduce the residual, we first obtain some good estimation of the group sparse coefficients of the original image by the first-pass estimation of noisy image, and then centralize the group sparse coefficients of noisy image to the estimation. Experimental results have demonstrated that the proposed method not only outperforms many state-of-the-art denoising methods such as BM3D and WNNM, but  results in a competitive speed.
\section{Modeling of Group sparse residual constraint}
\label{sec:format}
\subsection{Group-based sparse representation}
\label{ssec:subhead}
Recent studies \cite{8,13,14,15,24} have revealed that structured or group sparsity can offer more powerful reconstruction performance for image denoising. More specifically, image $\textbf{\emph{X}}$ with size $\emph{N}$ is divided into $\emph{n}$ overlapped patches $\textbf{\emph{x}}_i$ of size $\sqrt{bc}\times\sqrt{bc}, i=1,2,...,n$.  Then, for each $\textbf{\emph{x}}_i$, its most similar $k$ patches are selected from a $L \times L$ sized searching window to form a set ${\textbf{\emph{S}}}_i$. After this, all the patches in ${\textbf{\emph{S}}}_i$ are stacked into a matrix ${\textbf{\emph{X}}}_i\in\Re^{{bc}\times {k}}$, which contains every element of ${\textbf{\emph{S}}}_i$ as its column, i.e., ${\textbf{\emph{X}}}_i=\{{\textbf{\emph{x}}}_{i,1}, {\textbf{\emph{x}}}_{i,2}, ..., {\textbf{\emph{x}}}_{i,k}\}$. The matrix ${\textbf{\emph{X}}}_i$ consisting of  all the patches with similar structures is called as a group, where ${\textbf{\emph{x}}_{i,k}}$ denotes the $k$-th similar patch (column form) of the $i$-th group. Finally, similar to patch-based sparse representation \cite{5}, give a dictionary ${\textbf{\emph{D}}}_i$, which is often learned from each group. Thus, each group ${\textbf{\emph{X}}}_i$ can be sparsely represented as ${\textbf{\emph{B}}}_i={{\textbf{\emph{D}}}_i}^T\textbf{\emph{X}}_i$ and solved by the following $\ell_p$-norm minimization problem,
\begin{equation}
{{\textbf{\emph{B}}}_i}=\arg\min\nolimits_{{\textbf{\emph{B}}}_i} \{||{\textbf{\emph{X}}}_i-{\textbf{\emph{D}}}_i{{\textbf{\emph{B}}}_i}||_F^2+\lambda||{{\textbf{\emph{B}}}_i}||_p\}
\end{equation} 
where $\lambda$ is the regularization parameter, and $p$ characterizes the sparsity of ${{\textbf{\emph{B}}}_i}$. Then the whole image ${\textbf{\emph{X}}}$ can be represented by the set of group sparse codes ${{\textbf{\emph{B}}}_i}$.

Thus, in image denoising, the goal is to exploit group sparse-based model to recover ${\textbf{\emph{X}}}_i$ from ${\textbf{\emph{Y}}}_i$, and to solve the following minimization problem,
\begin{equation}
{{\textbf{\emph{A}}}_i}=\arg\min\nolimits_{{\textbf{\emph{A}}}_i} \{||{\textbf{\emph{Y}}}_i-{\textbf{\emph{D}}}_i{{\textbf{\emph{A}}}_i}||_F^2+\lambda||{{\textbf{\emph{A}}}_i}||_p\}
\end{equation} 

Once all group sparse codes ${{\textbf{\emph{A}}}_i}$ are achieved, the latent clean image ${{\textbf{\emph{X}}}}$ can be reconstructed as $\hat{\textbf{\emph{X}}}={\textbf{\emph{D}}}{\textbf{\emph{A}}}$.

Although group sparsity has demonstrated its effectiveness in image denoising, most  existing methods only use the NSS prior of noisy input images for noise removal (e.g., (2)), making it challenging to recover the latent clean image ${{\textbf{\emph{X}}}}$ directly from its noisy observation ${{\textbf{\emph{Y}}}}$.

\subsection{Group sparsity residual}
\label{ssec:subhead}
Let us revisit (1) and (2), due to the influence of noise, it is very difficult to estimate the true group sparse code ${{\textbf{\emph{B}}}}$ from noisy image ${{\textbf{\emph{Y}}}}$. In other words, the group sparse code ${{\textbf{\emph{A}}}}$ obtained by solving (2) are expected to be close enough to the true group sparse code ${{\textbf{\emph{B}}}}$ of the original image ${{\textbf{\emph{X}}}}$. As a consequence, the quality of image denoising largely depends on the level of the group sparsity residual,  which is defined as the difference between group sparse code ${{\textbf{\emph{A}}}}$ and true group sparse code ${{\textbf{\emph{B}}}}$,
\begin{equation}
{{\textbf{\emph{R}}}}={{\textbf{\emph{A}}}}-{{\textbf{\emph{B}}}}
\end{equation} 

Therefore, to reduce the group sparsity residual ${{\textbf{\emph{R}}}}$ and enhance the accuracy of  ${{\textbf{\emph{A}}}}$, we propose a new prior model to image denoising by group sparse residual constraint (GSRC), we can rewrite (2) into
\begin{equation}
{{\textbf{\emph{A}}}_i}=\arg\min\nolimits_{{\textbf{\emph{A}}}_i} \{||{\textbf{\emph{Y}}}_i-{\textbf{\emph{D}}}_i{{\textbf{\emph{A}}}_i}||_F^2+\lambda||{{\textbf{\emph{A}}}_i}-{{\textbf{\emph{B}}}_i}||_p\}
\end{equation} 

In practice, image ${{\textbf{\emph{X}}}}$ is not available. Therefore, we cannot obtain the true group sparse code ${{\textbf{\emph{B}}}}$. Nonetheless, we can compute some good estimation of ${{\textbf{\emph{B}}}}$. In general, there are various methods to estimate the true group sparse code ${{\textbf{\emph{B}}}}$, which  depend on the prior knowledge of ${{\textbf{\emph{B}}}}$ we have. For example, if we have many example images  are similar to the original image ${{\textbf{\emph{X}}}}$, then  a good estimation of ${{\textbf{\emph{B}}}}$ could be learned from the example image set. However, the example image set is simply and unsuitable under many real situations.

The strategy of $\emph{first-pass estimation}$ is a popular means to image denoising. The basic idea is as follows. Firstly, the original noise image was initially operated by using some denoising algorithms (e.g., BM3D \cite{13}, NLM \cite{7}, EPLL \cite{16}, etc.). Secondly, the result of $\emph{first-pass estimation}$  would continue to be denoised by the proposed method. In past two years, a variety of image denoising methods based on $\emph{first-pass estimation}$ have been developed, such as GID method \cite{17}, SOS method \cite{18}, TID method \cite{19}, MSEPLL method \cite{20}, etc.

Based on the above analysis, we first apply BM3D \cite{13} to noisy image ${{\textbf{\emph{Y}}}}$, and then the initialization result of BM3D is defined as ${{\textbf{\emph{Z}}}}$. Since the BM3D has an ideal denoising performance, ${{\textbf{\emph{Z}}}}$ could be regarded as a good approximation of the original image ${{\textbf{\emph{X}}}}$. Therefore, in this paper we can achieve the true group sparse code ${{\textbf{\emph{B}}}}$ from ${{\textbf{\emph{Z}}}}$.

Note that in this paper the noise image ${{\textbf{\emph{Y}}}}$ is only operated by BM3D, because  BM3D result is regarded as a good approximation of the original image ${{\textbf{\emph{X}}}}$, without continuing denoising operator to BM3D result.

\section{Algorithm of GSRC}
\label{sec:pagestyle}
In (4), except for estimating ${{\textbf{\emph{B}}}}$, we also need to determine the value of $p$. Here we perform some experiments to investigate the statistical property of ${{\textbf{\emph{R}}}}$, where ${{\textbf{\emph{R}}}}$ represents the set of ${{\textbf{\emph{R}}}}_i={{\textbf{\emph{A}}}}_i-{{\textbf{\emph{B}}}}_i$. In these experiments, an image $\emph{Leave}$ is used as an example in the case of image denoising, where the original image ${{\textbf{\emph{X}}}}$ is added by Gaussian white noise with standard deviation $\sigma$= 30. We plot the histogram of ${{\textbf{\emph{R}}}}$ as well as the fitting Gaussian, Laplacian and hyper-Laplacian distribution of ${{\textbf{\emph{R}}}}$ in Fig. 1(a). To better observe the fitting of the tails, we also plot these distributions in the log domain in Fig. 1(b). It can be seen that the histogram of ${{\textbf{\emph{R}}}}$ can be well characterized  by the Laplacian distribution. Thus, the $\ell_1$-norm is adopted to regularize ${{\textbf{\emph{R}}}_i}$, and (4) can be rewritten as
\begin{figure}[htb]
\begin{minipage}[b]{1\linewidth}
  \centering
  \centerline{\includegraphics[width=8cm]{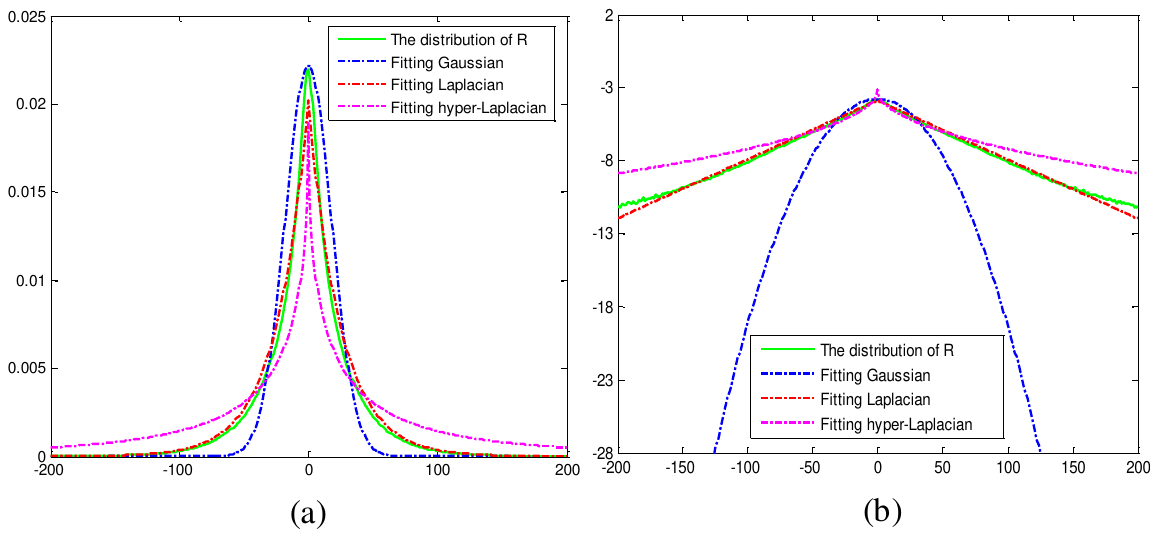}}
\end{minipage}
\caption{The distribution of ${{\textbf{\emph{R}}}}$ for image $\emph{Leave}$ with $\sigma$=30 and fitting Gaussian, Laplacian and hyper-Laplacian distribution in (a) linear and (b) log domain, respectively.}
\label{fig:1}
\end{figure}
\begin{equation}
\begin{aligned}
{{\textbf{\emph{A}}}_i}& = \arg\min\nolimits_{{\textbf{\emph{A}}}_i} \{||{\textbf{\emph{Y}}}_i-{\textbf{\emph{D}}}_i{{\textbf{\emph{A}}}_i}||_F^2+\lambda||{{\textbf{\emph{A}}}_i}-{{\textbf{\emph{B}}}_i}||_1\}\\
& =\arg\min\nolimits_{\tilde{\boldsymbol\alpha}_i} \{||\tilde{{\textbf{\emph{y}}}}_i-\tilde{{\textbf{\emph{D}}}}_i{\tilde{\boldsymbol\alpha}_i}||_2^2+\lambda||{\tilde{\boldsymbol\alpha}_i}-{\tilde{\boldsymbol\beta}_i}||_1\}
\label{eq:11}
\end{aligned}
\end{equation} 
where $\tilde{{\textbf{\emph{y}}}}_i, {\tilde{\boldsymbol\alpha}_i}$, and ${\tilde{\boldsymbol\beta}_i}$ denote the vectorization of the matrix ${{\textbf{\emph{Y}}}_i}, {{\textbf{\emph{A}}}_i}$ and ${{\textbf{\emph{B}}}_i}$, respectively. Each column $\tilde{{\textbf{\emph{d}}}}_h$ of the matrix $\tilde{{\textbf{\emph{D}}}}_i=[\tilde{{\textbf{\emph{d}}}}_1, \tilde{{\textbf{\emph{d}}}}_2, ..., \tilde{{\textbf{\emph{d}}}}_J]$  denotes the vectorization of the rank-one matrix, where $J$ denotes the number of dictionary atoms.

For fixed ${\tilde{\boldsymbol\beta}_i}, \lambda$, (5) is convex and can be solved efficiently. We adopt the surrogate algorithm in \cite{21} to solve (5). In the $t+1$-iteration, the proposed shrinkage operator can be calculated as
\begin{equation}
{\tilde{\boldsymbol\alpha}_i}^{t+1}= {{{\textbf{\emph{S}}}}_{\lambda}}({\tilde{{\textbf{\emph{D}}}}_i}^{-1}{{\hat{\tilde{{\textbf{\emph{x}}}}}}_i}^{t}-{{\tilde{\boldsymbol\beta}_i}})+{{\tilde{\boldsymbol\beta}_i}}
\label{eq:12}
\end{equation} 
where ${{{\textbf{\emph{S}}}}_{\lambda}}(\cdot)$ is the soft-thresholding operator, ${{\hat{\tilde{{\textbf{\emph{x}}}}}}_i}$  represents the vectorization of the $i$-th reconstructed group ${{\hat{{\textbf{\emph{X}}}}}_i}$. The above shrinkage operator follows the standard surrogate algorithm, from which more details can be seen in \cite{21}.

The parameter $\lambda$ that balances the fidelity term and the regularization term should be adaptively determined for better denoising performance. Inspired by \cite{1}, the regularization parameter $\lambda$ of each group ${\textbf{\emph{Y}}}_i$ is set as ${{{\lambda}}}={c*2\sqrt{2}{\sigma_n}^2}/{{{\sigma}}}$, where ${{\sigma}}$ denotes the estimated variance of ${\textbf{\emph{R}}}_i$, and $c$ is a small constant.

With the solution ${{\textbf{\emph{A}}}_i}$ in (6), the clean group ${{\textbf{\emph{X}}}_i}$ can be reconstructed as ${{\hat{\textbf{\emph{X}}}}_i}={{\textbf{\emph{D}}}_i}{{\textbf{\emph{A}}}_i}$. Then the clean image ${{\hat{\textbf{\emph{X}}}}}$ can be reconstructed by aggregating all the groups $\{{{\textbf{\emph{X}}}_i}\}$. In practical, we could perform the above denoising procedures for better results by several iterations. In the $t$+1-th iteration, the iterative regularization strategy \cite{22} is used to update the estimation of noise variance. Then the standard divation of noise in $t$+1-th iteration is adjusted as ${(\sigma^{t+1})}=\gamma*\sqrt{({\sigma^2-||{{\textbf{\emph{Y}}}}-{\hat{{\textbf{\emph{X}}}}}^{t+1}||_2^2})}$, where $\gamma$ is a constant.
$k$NN method has been widely used to similar patch selection. In order to obtain an effective similar patches index by $k$NN, we empirically define SSIM ( ${\hat{{\textbf{\emph{X}}}}}^{t+1}$, {{\textbf{\emph{Z}}}})-SSIM (${\hat{{\textbf{\emph{X}}}}}^{t}$, {{\textbf{\emph{Z}}}})$<$ $\tau$, then ${\hat{{\textbf{\emph{X}}}}}^{t+1}$ is regarded as target image to fetch the $k$ similar patches index, otherwise  ${{\textbf{\emph{Z}}}}$ is regarded as target image, where SSIM represents structural similarity, and $\tau$ is a small constant.
The complete description of the proposed method for image denoising via GSRC model is exhibited in $\textbf{Algorithm 1}$.
 \begin{table}[htb]
\centering  
\begin{tabular}{lccc}  
\hline  
\qquad \qquad \qquad \qquad Image denoising via GSRC\\
\hline
$\textbf{Input:}$ \ Noisy image ${{\textbf{\emph{Y}}}}$.\\

  $\rm \textbf{Initialization:} \  {\hat{{\textbf{\emph{X}}}}}={{\textbf{\emph{Y}}}}, {{\textbf{\emph{Z}}}}, \emph{c}, \emph{k}, \emph{bc}, \emph{L}, \sigma, \tau, \gamma, \delta$;\\
  $\rm \textbf{For}$\ $t=1, 2, ..., iter$ $\rm \textbf{do}$\\
  \qquad Iterative regularization ${{\textbf{\emph{Y}}}}^{t+1}= {{\textbf{\emph{X}}}}^{t}+\delta({{\textbf{\emph{Y}}}}-{{\textbf{\emph{X}}}}^{t})$;\\
  \qquad $\textbf{If}$\qquad ${\rm SSIM}({{\textbf{\emph{Y}}}}^{t+1}, {{\textbf{\emph{Z}}}})-{\rm SSIM}({{\textbf{\emph{Y}}}}^{t}, {{\textbf{\emph{Z}}}})<\tau$\\
  \qquad \qquad Similar patch index based on ${{\textbf{\emph{Y}}}}^{t+1}$.\\
  \qquad $\textbf{Else}$\\
   \qquad \qquad Similar patch index based on ${{\textbf{\emph{Z}}}}$.\\
   \qquad $\textbf{End if}$\\
 \qquad $\rm \textbf{For}$\ each patch ${{\textbf{\emph{y}}}}_i$ and ${{\textbf{\emph{z}}}}_i$ $\rm \textbf{do}$\\
   \qquad \qquad  Find a group ${{{\textbf{\emph{Y}}}}_i}^{t+1}$ via $k$NN.\\
   \qquad \qquad Find a group ${{{\textbf{\emph{Z}}}}_i}^{t+1}$ via $k$NN.\\
   \qquad \qquad Constructing dictionary ${{{\textbf{\emph{D}}}}_i}^{t+1}$ by ${{\textbf{\emph{Y}}}}_i$ using PCA operator.\\
   \qquad \qquad Update ${{{\textbf{\emph{B}}}}_i}^{t+1}$ computing by ${{\textbf{\emph{B}}}}_i={{{\textbf{\emph{D}}}}_i}^{-1}{{\textbf{\emph{Z}}}}_i$.\\
   \qquad \qquad Update ${\lambda}^{t+1}$ computing by ${{{\lambda}}}={c*2\sqrt{2}{\sigma_n}^2}/{{{\sigma}}}$.\\
   \qquad \qquad Update ${{{\textbf{\emph{A}}}}_i}^{t+1}$ computing by (6).\\
   \qquad \qquad Get the estimation ${{{\textbf{\emph{X}}}}_i}^{t+1}$ =${{{\textbf{\emph{D}}}}_i}^{t+1}$${{{\textbf{\emph{A}}}}_i}^{t+1}$.\\
  \qquad  $\rm \textbf{End for}$\\
   \qquad \qquad Aggregate ${{{\textbf{\emph{X}}}}_i}^{t+1}$ to form the recovered image ${\hat{{\textbf{\emph{X}}}}}^{t+1}$.\\
    $\rm \textbf{End for}$\\
     $\textbf{Output:}$ ${\hat{\textbf{\emph{X}}}}^{t+1}$.\\
\hline
\end{tabular}
\end{table}
\begin{table*}[htb]
\caption{Denoising PSNR (dB) results by different denoising methods}
\scriptsize
\centering  
\begin{tabular}{|c||c|c|c|c|c|c||c|c|c|c|c|c|c|}
\hline
  \multicolumn{1}{|c||}{}&\multicolumn{6}{|c||}{$\sigma=30$}&\multicolumn{6}{|c|}{$\sigma=40$}\\
\hline\hline
\multirow{1}{*}{\textbf{{Images}}}&{\textbf{{BM3D}}}&{\textbf{{NCSR}}}&{\textbf{{WNNM}}}
&{\textbf{{AST-NLS}}}&{\textbf{{MSEPLL}}}&{\textbf{{GSRC}}}&{\textbf{{BM3D}}}
&{\textbf{{NCSR}}}&{\textbf{{WNNM}}}&{\textbf{{AST-NLS}}}&{\textbf{{MSEPLL}}}&{\textbf{{GSRC}}}\\
 \hline
 \multirow{1}{*}{Barbara}& 29.18 &  28.90  & 29.56  & 29.30  & 27.72  & \textbf{29.65} & 27.33 & 27.36 & 27.84 & 27.51 & 26.04 & \textbf{28.00}\\
 \hline
 \multirow{1}{*}{C.Man}  & 28.68 &  28.59  & \textbf{28.81}  & 28.72  & 28.37 & 28.79 & 27.20 & 27.12 & 27.58 & 27.41 & 27.08 & \textbf{27.63}\\
  \hline
 \multirow{1}{*}{Couple} & 30.58 &  30.22 & 30.56  & 30.30  & 30.42  & \textbf{30.67} & 29.13 & 28.89 & 29.24 & 28.89  & 29.15 & \textbf{29.38}\\
  \hline
 \multirow{1}{*}{foreman}& 32.69 &  32.78 & 33.25  & 32.87  & 32.34  & \textbf{33.37} & 31.11 & 31.59 & 31.81 & 31.27  & 31.05 & \textbf{32.05}\\
  \hline
 \multirow{1}{*}{girl}   & 30.71 &  30.57 & 30.72  & 30.58  & \textbf{30.79} & 30.79  & 29.64 & 29.65 & 29.68 & 29.38  & \textbf{29.80} & 29.79\\
  \hline
 \multirow{1}{*}{Hill}   & 28.44 &  28.25 & 28.58  & 28.37  & 28.41  & \textbf{28.63} & 27.21 & 27.02 & 27.25 & 27.08  & 27.18 & \textbf{27.31}\\
  \hline
 \multirow{1}{*}{House}  & 32.08 &  32.08 & 32.58  & 32.49  & 31.71  & \textbf{32.66} & 30.67 & 30.82 & 31.35 & 31.16  & 30.47 & \textbf{31.64}\\
  \hline
 \multirow{1}{*}{Leave}  & 27.82 &  28.17 & 28.65  & 28.46  & 27.26  & \textbf{28.79} & 25.69 & 26.24 & \textbf{27.00} & 26.86 & 25.72 & 26.96\\
  \hline
 \multirow{1}{*}{Lena}   & 29.55 &  29.43 & 29.83  & 29.52  & 29.46  & \textbf{29.87} & 27.77 & 28.01 & 28.38 & 28.12  & 28.05 & \textbf{28.41}\\
  \hline
 \multirow{1}{*}{lin}    & 31.07 &  30.84 & 30.96  & 30.84  & 30.96  & \textbf{31.21} & 29.53 & 29.45 & 29.80 & 29.44  & 29.68 & \textbf{29.98}\\
 \hline
 \multirow{1}{*}{Monarch}& 28.39 &  28.47 & 29.02  & 28.73  & 28.49  & \textbf{29.03} & 26.74 & 26.86 & 27.54 & 27.30  & 27.06 & \textbf{27.59}\\
 \hline
 \multirow{1}{*}{Parrot} & 30.33 &  30.38 & 30.66  & 30.52  & 30.29  & \textbf{30.79} & 28.59 & 28.96 & 29.38 & 29.03  & 28.94 & \textbf{29.58}\\
 \hline
 \multirow{1}{*}{\textbf{Average}}& 29.96 &  29.89 & 30.27  & 30.06  & 29.68  & \textbf{30.35} & 28.38 & 28.50 & 28.91 & 28.62  & 28.35 & \textbf{29.03}\\
 \hline
\hline\hline
  \multicolumn{1}{|c||}{}&\multicolumn{6}{|c||}{$\sigma=50$}&\multicolumn{6}{|c|}{$\sigma=100$}\\
\hline\hline
\multirow{1}{*}{\textbf{{Images}}}&{\textbf{{BM3D}}}&{\textbf{{NCSR}}}&{\textbf{{WNNM}}}
&{\textbf{{AST-NLS}}}&{\textbf{{MSEPLL}}}&{\textbf{{GSRC}}}&{\textbf{{BM3D}}}
&{\textbf{{NCSR}}}&{\textbf{{WNNM}}}&{\textbf{{AST-NLS}}}&{\textbf{{MSEPLL}}}&{\textbf{{GSRC}}}\\
 \hline
 \multirow{1}{*}{Barbara}  & 26.45  &  26.25  & 26.69  & 26.41 & 25.06 & \textbf{26.79} & 23.05 & 22.84 & 23.44 & 23.11 & 22.11 & \textbf{23.47}\\
 \hline
 \multirow{1}{*}{C.Man}    & 26.22  &  26.17  & 26.45  & 26.34 & 26.12 & \textbf{26.71} & 23.09 & 22.94 & 23.40 & 23.34 & 23.01 & \textbf{23.60}\\
  \hline
 \multirow{1}{*}{Couple}   & 28.18  &  27.80  & 28.18  & 27.90 & 28.23 & \textbf{28.31} & 25.32 & 24.79 & 25.30 & 24.89 & 25.27 & \textbf{25.35}\\
  \hline
 \multirow{1}{*}{foreman}  & 30.24  &  30.56  & \textbf{31.25} & 30.60 & 30.04 & 31.12  & 26.31 & 26.55 & 27.40 & 27.06 & 26.84 & \textbf{27.55}\\
  \hline
 \multirow{1}{*}{girl}     & 28.91  &  28.83  & 28.94  & 28.69 & \textbf{29.07} & 29.01 & 26.17 & 26.30 & 26.43 & 26.17 & \textbf{26.89} & 26.69\\
  \hline
 \multirow{1}{*}{Hill}     & 26.27  &  26.06  & \textbf{26.45} & 26.22 & 26.28  & 26.40 & 23.63 & 23.41 & 23.76 & 23.28 & 23.74 & \textbf{23.91}\\
  \hline
 \multirow{1}{*}{House}    & 29.75  &  29.69  & 30.40  & 30.22 & 29.47 & \textbf{30.64} & 25.87 & 25.57 & 26.72 & 26.92 & 25.99 & \textbf{27.24}\\
  \hline
 \multirow{1}{*}{Leave}    & 24.75  &  24.97  & 25.49  & 25.51 & 24.42 & \textbf{25.75} & 20.94 & 20.90 & \textbf{21.58} & 21.45 & 20.31 & 21.55\\
  \hline
 \multirow{1}{*}{Lena}     & 27.03  &  27.02  & 27.16  & 27.08 & 26.97 & \textbf{27.28} & 24.01 & 23.70 & 24.42 & 24.10  & 23.91 & \textbf{24.43}\\
  \hline
 \multirow{1}{*}{lin}      & 28.75  &  28.40  & \textbf{28.90} & 28.52 & 28.69 & 28.90  & 25.66 & 25.22 & \textbf{25.90} & 25.67 & 25.63 & 25.89\\
  \hline
 \multirow{1}{*}{Monarch}  & 25.89  &  25.78  & 26.42 & 26.10  & 25.93 & \textbf{26.64} & 22.55 & 22.14 & 22.95 & 22.70  & 22.44 & \textbf{23.03}\\
   \hline
 \multirow{1}{*}{Parrot}   & 28.06  &  27.88  & 28.30 & 28.06  & 27.90 & \textbf{28.51} & 24.62 & 24.46 & 24.94 & 24.88  & 24.38 & \textbf{25.23}\\
   \hline
 \multirow{1}{*}{\textbf{Average}}  & 27.54   & 27.45 & 27.89  & 27.64 & 27.35 & \textbf{28.01} & 24.27 & 24.06 & 24.69  & 24.46 & 24.21 & \textbf{24.83}\\
 \hline
\end{tabular}
\end{table*}
\begin{figure}[!htbp]
\begin{minipage}[b]{1\linewidth}
  \centering
  \centerline{\includegraphics[width=7.5cm]{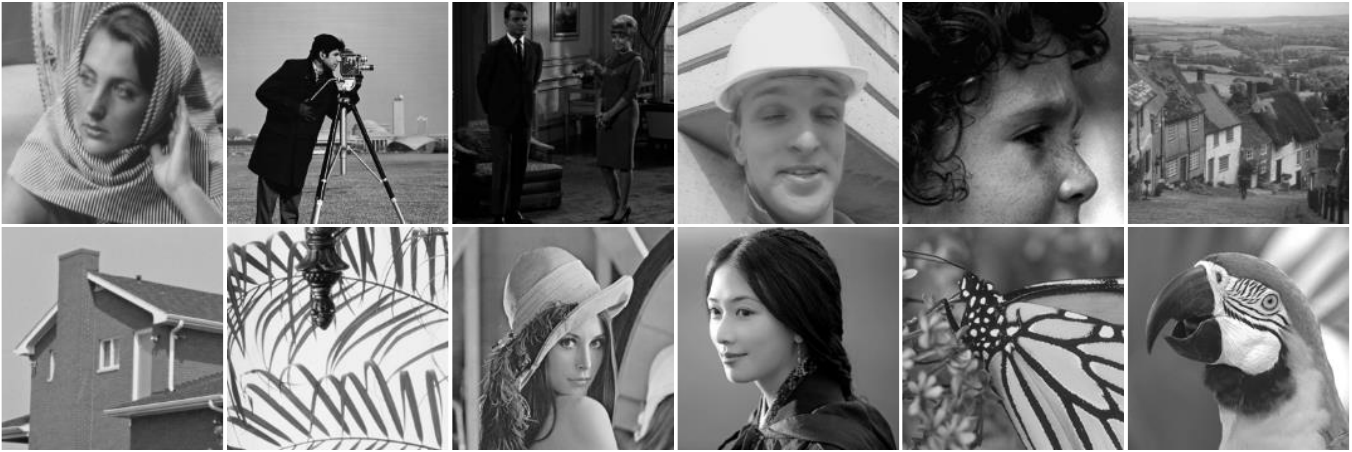}}
\end{minipage}
\caption{All test images.}
\label{fig:2}
\end{figure}
\section{Experimental Results}
\label{sec:typestyle}
In this section, we validate the performance of the proposed method and compare it with recently proposed state-of-the-art denoising methods, including BM3D \cite{13}, NCSR \cite{6}, WNNM \cite{15}, AST-NLS \cite{23}, and MSEPLL \cite{20}. The parameter setting of GSRC is as follows: the searching window $L \times L$ for similar patches is set to be  $30\times30$ and $\tau$ is 0.0001. The size of each patch is set to be $6\times 6$, $7\times 7$, $8\times 8$ and $9\times 9$ for $\sigma\leq20$, $20<\sigma\leq50$, $50<\sigma\leq75$ and $75<\sigma\leq100$, respectively. $(c, \gamma, \delta)$ are set to (0.2, 0.18, 0.67) when $\sigma\leq30$ or  (0.3, 0.22, 0.67) otherwise. The source code of the proposed can be downloaded at: \textcolor{red}{\url{http://www.escience.cn/system/file?fileId=85492}}.

We first evaluate the competing methods on 12 test images, whose scenes are shown in Fig. 2. Gaussian white noise with standard deviation $\sigma=30, 40, 50, 100$ is added to those  test images. The PSNR results by the competing denoising methods are shown in Table 1. It can be seen that the proposed GSRC performed competitively compared to other methods. It achieves 0.39-0.56dB, 0.46-0.76dB, 0.08-0.14dB, 0.3-0.41dB and 0.61-0.66dB improvement on average are the BM3D, NCSR, WNNM, AST-NLS and MSEPLL, respectively. The proposed GSRC can obtain better denoising results in the case of strong noise corruption. For example, in the case of $\sigma=100$, the PSNR gains of the GSRC over the benchmark BM3D method can be as much 1.24dB ad 1.37dB on $\emph{foreman}$ and $\emph{House}$, respectively. The visual comparison of competing denoising methods at noise level 40 and 100 are shown in Fig. 3 and Fig. 4, respectively. It can be found out that BM3D, NCSR, WNNM, AST-NLS and MSEPLL still generated some undesirable artifacts and some details are lost. By contrast, the proposed GSRC is able to preserve the sharp edges and suppress undesirable artifacts more effectively than the other competing methods. Such experimental findings clearly demonstrate that the GSRC  model is a stronger prior for the class of photographic images containing large variations in edges/textures.
\begin{figure}[htb]
\begin{minipage}[b]{1\linewidth}
  \centering
  \centerline{\includegraphics[width=7.5cm]{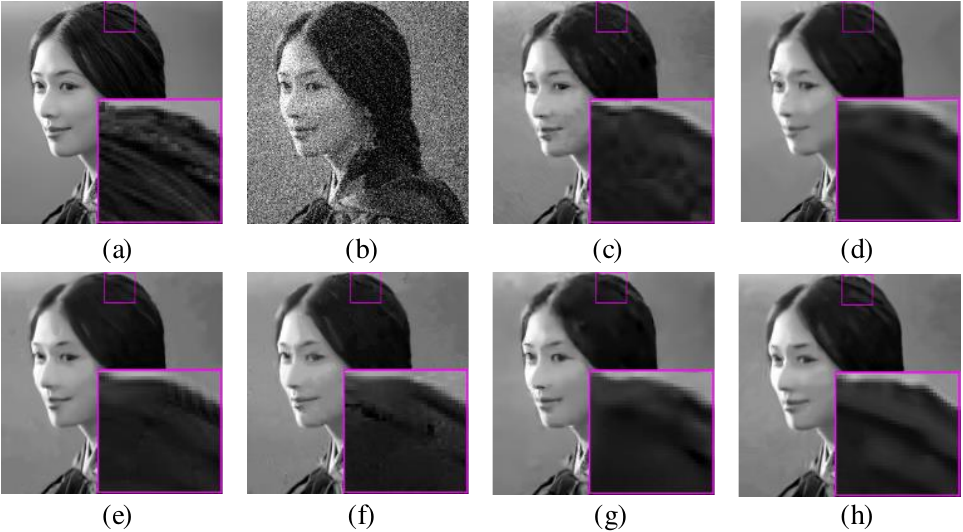}}
\end{minipage}
\caption{Denoising images of ${lin}$ by different methods ($\sigma=40$). (a) Original image; (b) Noisy image; (c) BM3D (PSNR=29.53dB); (d) NCSR (PSNR=29.45dB); (e) WNNM (PSNR=29.80dB); (f) AST-NLS (PSNR=29.43dB); (g) MSEPLL (PSNR=29.68dB); (h) GSRC (PSNR=\textbf{29.98dB}).}
\label{fig:3}
\end{figure}
\begin{figure}[htb]
\begin{minipage}[b]{1\linewidth}
  \centering
  \centerline{\includegraphics[width=7.5cm]{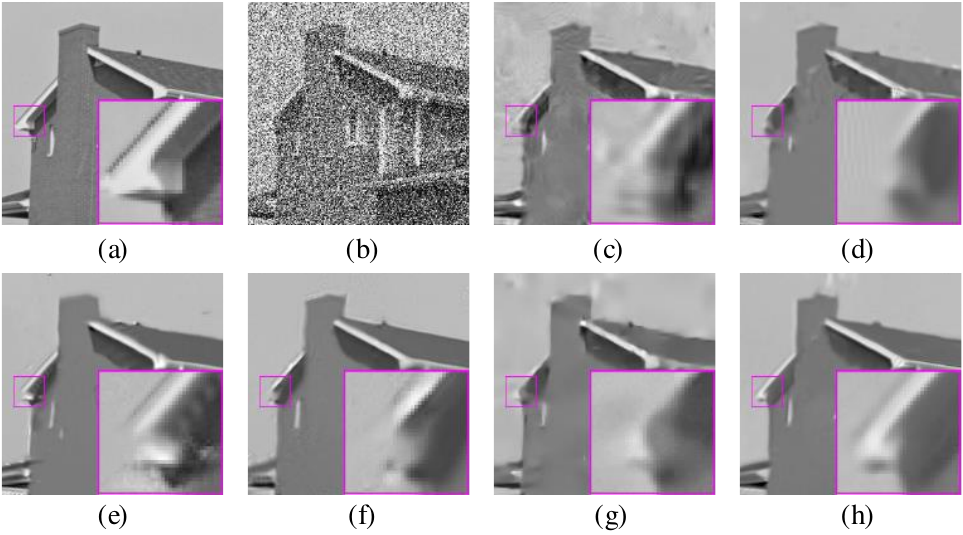}}
\end{minipage}
\caption{Denoising images of ${House}$ by different methods ($\sigma=100$). (a) Original image; (b) Noisy image; (c) BM3D (PSNR=25.87dB); (d) NCSR (PSNR=25.57dB); (e) WNNM (PSNR=26.72dB); (f) AST-NLS (PSNR=26.92dB); (g) MSEPLL (PSNR=25.99dB); (h) GSRC (PSNR=\textbf{27.24dB}).}
\label{fig:4}
\end{figure}
\begin{table*}[htb]
\caption{Average run time (seconds) with standard deviation of different methods on test images. BM3D is implemented with complied C++ mex-function and uses parallelization, while the other methods are implemented in Matlab.}
\scriptsize
\centering  
\begin{tabular}{|c|c|c|c|c|c|c|}
\hline\hline
\multirow{1}{*}{\textbf{{$\sigma$}}}&{\textbf{{BM3D}}}&{\textbf{{NCSR}}}&{\textbf{{WNNM}}}
&{\textbf{{AST}}}&{\textbf{{MSEPLL}}}&{\textbf{{GSRC}}}\\
 \hline
 \multirow{1}{*}{30}& 2.01$\pm$0.04 &  211.76$\pm$6.41 & 207.50$\pm$1.44   & 295.42$\pm$2.55   & 178.07$\pm$3.40   & 67.51$\pm$11.05\\
 \hline
 \multirow{1}{*}{40}& 1.91$\pm$0.04 &  467.35$\pm$17.99 & 207.75$\pm$1.77  & 295.14$\pm$2.41   & 178.01$\pm$1.56   & 55.20$\pm$7.72\\
  \hline
 \multirow{1}{*}{50}& 4.58$\pm$0.06 &  464.03$\pm$15.56 & 159.63$\pm$1.12  & 494.22$\pm$3.87   & 255.85$\pm$2.07   & 54.70$\pm$6.44\\
  \hline
 \multirow{1}{*}{100}& 4.81$\pm$0.06 &  345.08$\pm$11.43 & 256.00$\pm$1.50 & 1118.84$\pm$4.79   & 256.70$\pm$2.32   & 114.64$\pm$13.85\\
 \hline\hline
\end{tabular}
\end{table*}

Efficiency is another key factor in evaluating an algorithm. We then compare the speed of all competing methods. All experiments are conducted under the Matlab 2012b environment on a machine with Intel (R) Core (TM) i3-4150 with 3.56Hz CPU and 4GB memory. The run time of the competing methods on the test images is shown in Table 2. It can be seen that the proposed GSRC used less computation time than all the other methods except BM3D. It may be because that BM3D was implemented with C++ mex-function and parallelization, while GSRC implemented purely in Matlab.
\section{Conclusion}
\label{sec:majhead}

In this paper, we propose a new prior model for image denoising via group sparsity residual constraint. To improve the performance of the group sparse-based image denoising, the concept of group sparsity residual is proposed, and thus, the problem of image denoising is transformed into one that reduces the group sparsity residual. To this end, we first obtain some good estimation of the group sparse coefficients of the original image by the first-pass estimation of noisy image, and then centralize the group sparse coefficients of noisy image to the estimation. Experimental results have demonstrated that the proposed method not only outperforms many state-of-the-art denoising methods such as BM3D and WNNM in terms of PSNR, but also results in a competitive time.


\begin{thebibliography}{99}
\bibitem{1}
Chang S G, Yu B, Vetterli M. Adaptive wavelet thresholding for image denoising and compression[J]. IEEE transactions on image processing, 2000, 9(9): 1532-1546.
\bibitem{2}
Remenyi N, Nicolis O, Nason G, et al. Image denoising with 2D scale-mixing complex wavelet transforms[J]. IEEE Transactions on Image Processing, 2014, 23(12): 5165-5174.
\bibitem{3}
Rudin L I, Osher S, Fatemi E. Nonlinear total variation based noise removal algorithms[J]. Physica D: Nonlinear Phenomena, 1992, 60(1): 259-268.
\bibitem{4}
Chambolle A. An algorithm for total variation minimization and applications[J]. Journal of Mathematical imaging and vision, 2004, 20(1-2): 89-97.
\bibitem{5}
Elad M, Aharon M. Image denoising via sparse and redundant representations over learned dictionaries[J]. IEEE Transactions on Image processing, 2006, 15(12): 3736-3745.
\bibitem{6}
Dong W, Zhang L, Shi G, et al. Nonlocally centralized sparse representation for image restoration[J]. IEEE Transactions on Image Processing, 2013, 22(4): 1620-1630.
\bibitem{7}
Buades A, Coll B, Morel J M. A non-local algorithm for image denoising[C]//2005 IEEE Computer Society Conference on Computer Vision and Pattern Recognition (CVPR'05). IEEE, 2005, 2: 60-65.
\bibitem{8}
Mairal J, Bach F, Ponce J, et al. Non-local sparse models for image restoration[C]//2009 IEEE 12th International Conference on Computer Vision. IEEE, 2009: 2272-2279.
\bibitem{9}
Gilboa G, Osher S. Nonlocal operators with applications to image processing[J]. Multiscale Modeling \& Simulation, 2008, 7(3): 1005-1028.
\bibitem{10}
Protter M, Elad M, Takeda H, et al. Generalizing the nonlocal-means to super-resolution reconstruction[J]. IEEE Transactions on image processing, 2009, 18(1): 36-51.
\bibitem{11}
Peyré G. Image processing with nonlocal spectral bases[J]. Multiscale Modeling \& Simulation, 2008, 7(2): 703-730.
\bibitem{12}
Zhang X, Burger M, Bresson X, et al. Bregmanized nonlocal regularization for deconvolution and sparse reconstruction[J]. SIAM Journal on Imaging Sciences, 2010, 3(3): 253-276.
\bibitem{13}
Dabov K, Foi A, Katkovnik V, et al. Image denoising by sparse 3-D transform-domain collaborative filtering[J]. IEEE Transactions on image processing, 2007, 16(8): 2080-2095.
\bibitem{14}
Ji H, Huang S, Shen Z, et al. Robust video restoration by joint sparse and low rank matrix approximation[J]. SIAM Journal on Imaging Sciences, 2011, 4(4): 1122-1142.
\bibitem{15}
Gu S, Zhang L, Zuo W, et al. Weighted nuclear norm minimization with application to image denoising[C]//Proceedings of the IEEE Conference on Computer Vision and Pattern Recognition. 2014: 2862-2869.
\bibitem{16}
Zoran D, Weiss Y. From learning models of natural image patches to whole image restoration[C]//2011 International Conference on Computer Vision. IEEE, 2011: 479-486.
\bibitem{17}
Talebi H, Milanfar P. Global image denoising[J]. IEEE Transactions on Image Processing, 2014, 23(2): 755-768.
\bibitem{18}
Romano Y, Elad M. Boosting of image denoising algorithms[J]. SIAM Journal on Imaging Sciences, 2015, 8(2): 1187-1219.
\bibitem{19}
Luo E, Chan S H, Nguyen T Q. Adaptive image denoising by targeted databases[J]. IEEE Transactions on Image Processing, 2015, 24(7): 2167-2181.
\bibitem{20}
Papyan V, Elad M. Multi-scale patch-based image restoration[J]. IEEE Transactions on Image Processing, 2016, 25(1): 249-261.
\bibitem{21}
Daubechies I, Defrise M, De Mol C. An iterative thresholding algorithm for linear inverse problems with a sparsity constraint[J]. Communications on pure and applied mathematics, 2004, 57(11): 1413-1457.
\bibitem{22}
Osher S, Burger M, Goldfarb D, et al. An iterative regularization method for total variation-based image restoration[J]. Multiscale Modeling \& Simulation, 2005, 4(2): 460-489.
\bibitem{23}
Liu H, Xiong R, Zhang J, et al. Image denoising via adaptive soft-thresholding based on non-local samples[C]//Proceedings of the IEEE Conference on Computer Vision and Pattern Recognition. 2015: 484-492.
\bibitem{24}
Li X, He H, Wang R, et al. Single image superresolution via directional group sparsity and directional features[J]. IEEE Transactions on Image Processing, 2015, 24(9): 2874-2888.
\bibitem{25}
Wan L, Han G, Shu L, et al. PD source diagnosis and localization in industrial high-voltage insulation system via multimodal joint sparse representation[J]. IEEE Transactions on Industrial Electronics, 2016, 63(4): 2506-2516.
\end{thebibliography}
\end{document}